\documentclass[letterpaper, 10 pt, conference]{ieeeconf}  

\IEEEoverridecommandlockouts                              

\usepackage{amsmath, amsfonts,bm}
\usepackage{graphicx}
\let\labelindent\relax
\usepackage{enumitem}
\usepackage{wrapfig}
\usepackage{comment}
\usepackage[hidelinks]{hyperref}

\def\Real{\mathbb{R}}
\renewcommand\Pr{{\operatorname{Pr}}}

\newcommand\deq{\overset{\mathrm{def}}{=\joinrel=}}
\def\ov{{\bf o}}
\def\qv{{\bf q}}
\def\yv{{\bf y}}
\def\zv{{\bf z}}
\def\clip{\operatorname{clip}}

\title{\LARGE \bf
Setting up a Reinforcement Learning Task with a Real-World Robot
}

\author{A. Rupam Mahmood$^{1}$, Dmytro Korenkevych$^{1}$, Brent J. Komer$^{2}$, and James Bergstra$^{1}$\\
{\tt\small \{rupam, dmytro.korenkevych\}@kindred.ai, bjkomer@uwaterloo.ca, james@kindred.ai}
\thanks{$^{1}$The authors are with Kindred Inc.}
\thanks{$^{2}$The author contributed during his internship at Kindred Inc.       
}%
}

\begin{document}

\maketitle
\thispagestyle{empty}
\pagestyle{empty}

\begin{abstract}
Reinforcement learning is a promising approach to developing hard-to-engineer adaptive solutions for complex and diverse robotic tasks.
However, learning with real-world robots is often unreliable and difficult, which resulted in their low adoption in reinforcement learning research.
This difficulty is worsened by the lack of guidelines for setting up learning tasks with robots.
In this work, we develop a learning task with a UR5 robotic arm to bring to light some key elements of a task setup and study their contributions to the challenges with robots.
We find that learning performance can be highly sensitive to the setup, and thus oversights and omissions in setup details can make effective learning, reproducibility, and fair comparison hard.
Our study suggests some mitigating steps to help future experimenters avoid difficulties and pitfalls.
We show that highly reliable and repeatable experiments can be performed in our setup, indicating the possibility of reinforcement learning research extensively based on real-world robots.
\end{abstract}

\section{Introduction}

Despite some recent successes (e.g., Levine et~al.~2016, Gu et al.\ 2017),
real-world robots are under-utilized in the quest for {general} reinforcement learning (RL) agents, which at this time is \mbox{primarily} confined to simulation.
This under-utilization is largely due to frustrations around unreliable and poor learning performance with robots. 
Although several RL methods are recently shown to be highly effective in simulations (Duan et al. 2016), they often yield poor performance when applied off-the-shelf to real-world tasks. 
Such \mbox{ineffectiveness} is sometimes attributed to some of the \mbox{integral} aspects of the real world including slow rate of data collection, partial observability, noisy sensors, safety, and frailty of physical devices.
This barrier contributed to a reliance on indirect approaches such as simulation-to-reality transfer (Rusu et al. 2017) and collective learning (Yahya et al. 2017, Gu et~al. 2017), which sometimes compensate for failures to learn from a single stream of real experience.

One oft-ignored shortcoming in real-world RL research is the lack of benchmark learning tasks or standards for setting up experiments with robots. 
Experiments with simulated robots are typically done on benchmark tasks with \mbox{easily} {available} simulators and {standardized} interfaces, relieving experimenters of many task-setup details such as the action space, the action cycle time and system delays.
On the other hand, setting up a learning task with real-world robots is far from obvious. 
An experimenter has to put a lot of work into establishing a device-specific sensorimotor interface between the learning agent and the robot as well as determining all the aspects of the environment that define the learning task. 
Such choices can be crucial for effective and reproducible learning performance. 
Unfortunately, RL research works with real-world robots typically do not describe many of these details, let alone study their effects in a controlled manner, although some notable exceptions exist (e.g., Schuitema et al.\ 2010, Degris et al.\ 2012, Hester \& Stone 2013).

\begin{figure}
\center
\includegraphics[scale=.55]{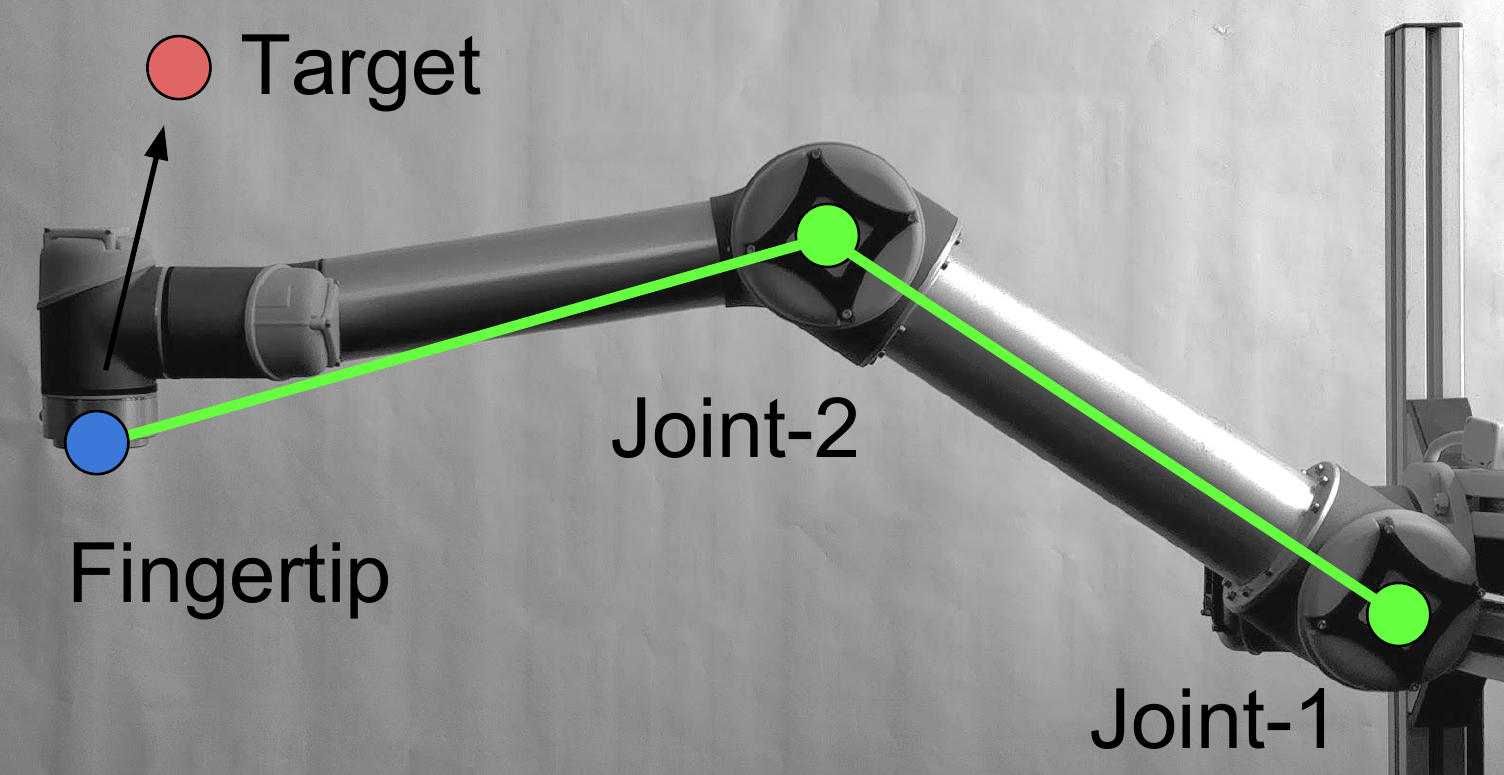}
\caption{
We use a UR5 robotic arm to define reinforcement learning tasks based on two joints (shown above) and six joints. The reward is the negative distance between the fingertip (blue) and a random target location (red).
}
\label{fig:reacher}
\end{figure}

In this work, we address the question of \emph{how to set up a real-world robotic task so that an off-the-shelf implementation of a standard RL method can perform effectively and reliably.}
We address this by developing a \emph{Reacher} task for the UR5 robotic arm (see Figure 1),
in which an agent learns to reach arbitrary target positions with low-level actuations  of a robotic arm using trial and error.
This task is easy to solve in a simulation but can be difficult with real-world robots (Gu et al.\ 2017). For the learning method,  we use the rllab implementation of TRPO (\mbox{Schulman et\! al.\! 2015}, Duan et\! al.\!~2016), a popular learning method with robust performance with respect to its hyper-parameters.

As we set up UR5 Reacher, we describe the steps and elements of setting up real-world RL tasks including the medium of data transmission, concurrency, {ordering} and delays of computations, low-level {actuation} types, and frequencies of operating them.
By {exploring} {different} variations of these elements, we study their {individual} contributions to the difficulty with robot learning.
We find that variability in time delays and choosing an action representation that is non-trivially processed before actuation can be highly detrimental to learning.
By accounting for these effects, we show that it is possible to achieve not only effective performance with real-world robots but also repeatability of learning from scratch in a highly reliable manner even when the repeats are run for hours on different times using different physical robots.

\section{Reinforcement Learning Task Formulation}

A reinforcement learning (RL) task is composed of an agent and an environment interacting with each other (Sutton \& Barto 2017), modeled formally as a \emph{Markov decision process} (MDP). 
In an MDP, an agent interacts with its environment at discrete time steps $t=1, 2, 3, \cdots$, where at each step $t$, the agent receives the environment's state information $S_t \in \cal{S}$ and a scalar reward signal $R_{t} \in \Real$. 
The agent uses a stochastic policy $\pi$ governed by a probability distribution $\pi(a|s) \deq \Pr\left\{ A_t=a | S_t=s \right\}$ to select an action $A_t \in \cal{A}$. 
The environment transitions to a new state $S_{t+1}$ and produces a new reward $R_{t+1}$ at the next time step using a transition probability distribution: $p(s', r|s, a) \deq \Pr\left\{ S_{t+1} = s', R_{t+1}=r | S_t = s, A_t = a \right\}$. 
The agent's performance is typically evaluated in terms of its future  accumulated rewards, known as a \mbox{return} $G_t \deq \sum_{k=t} \gamma^{k-t} R_{k+1}$, where $\gamma\in[0,1]$ is a discount factor.

The goal of the agent is typically to find a policy that maximizes the expected return. 
Such policies are often learned by estimating action values as in Q-learning or by directly parameterizing the policy and optimizing the policy parameters as in TRPO. 
In practice, the agent does not receive the environment's full state information but rather observes it partially through a real-valued observation vector $\ov_t$. 
Under this framework, an RL task is described primarily using three elements: the observation space, the action space, and the reward function.

\section{The UR5 Reacher Task}

In this section, we design a Reacher task with the UR5 robot, which we call \emph{UR5 Reacher}. 
We design it to be similar to OpenAI-Gym Reacher (Brockman et al.\ 2016), where an agent learns to reach arbitrary target positions with direct torque control of a simulated two-joint robotic arm.
By parameterizing the policy nonlinearly with a neural network, a policy-search method such as TRPO can solve Gym Reacher reasonably well in a few thousand time steps.
Designing the task based on Gym Reacher allows us to set a reasonable expectation of the learning time, utilize the choices already made, and isolate challenges that emerge from design decisions in the hardware interface.
In the following, we describe the interface of the UR5 robot and the details of the UR5 Reacher task.

The UR5 is a lightweight, flexible industrial robot with six joints manufactured by Universal Robots. The low-level robot controller of UR5, called \emph{URControl}, can be programmed by communicating over a TCP/IP connection. 
The robot can be controlled at the script-level using a programming language called  \emph{URScript}.
After establishing a connection, we can send URScript programs from a computer to URControl as strings over the socket. 
URScript programs run on URControl in real-time, which streams status packets every 8ms.
Each packet from the URControl contains the sensorimotor information of the robot including angular positions, velocities, target accelerations, and currents for all joints. 
The robot can be controlled with URScript by sending low-level actuation commands on the same 8ms clock.
The URScript \emph{servoj} command offers a position control interface, and \emph{speedj} offers a velocity control interface.
Unlike Gym Reacher, there is no torque control interface.

In  UR5 Reacher, we actuate the second and the third joints from the base.
We also extend it to a task with all six joints actuated, which we called \emph{UR5 Reacher 6D}.
The observation vector includes the joint angles, the joint velocities, and the vector difference between the target and the fingertip coordinates. 
Unlike Gym Reacher, we do not include the sines or cosines of joint angles or the target-position coordinates to simplify and reduce the observation space without losing essential information.
We include the previous action as part of the observation vector, which can be helpful for learning in systems with delays (Katsikopoulos \& Engelbrecht 2003). 
In Gym Reacher, the reward function is defined as: $R_{t} = -d_{t} - p_{t-1}$, where $d_t$ is the Euclidean distance between the target and the fingertip positions, and $p_t$ is the L2-norm of $A_t$ to penalize large torques. 
We use the same reward function but simplify it by dropping the penalty term.  
UR5 Reacher consists of episodes of interactions, where each episode is 4 seconds long to allow adequate exploration. 
The fingertip of UR5 is confined within a 2-dimensional $0.7m\times 0.5m$ boundary in UR5 Reacher and within a 3-dimensional $0.7m\times 0.5m\times 0.4m$ boundary in UR5 Reacher 6D.
At each episode, the target position is chosen randomly within the boundary, and the arm starts from the middle of the boundary.
In addition to constraining the fingertip within a boundary, the robot is also constrained within a joint-angular boundary to avoid self-collision.

\begin{figure*}
\center
\includegraphics[scale=.58]{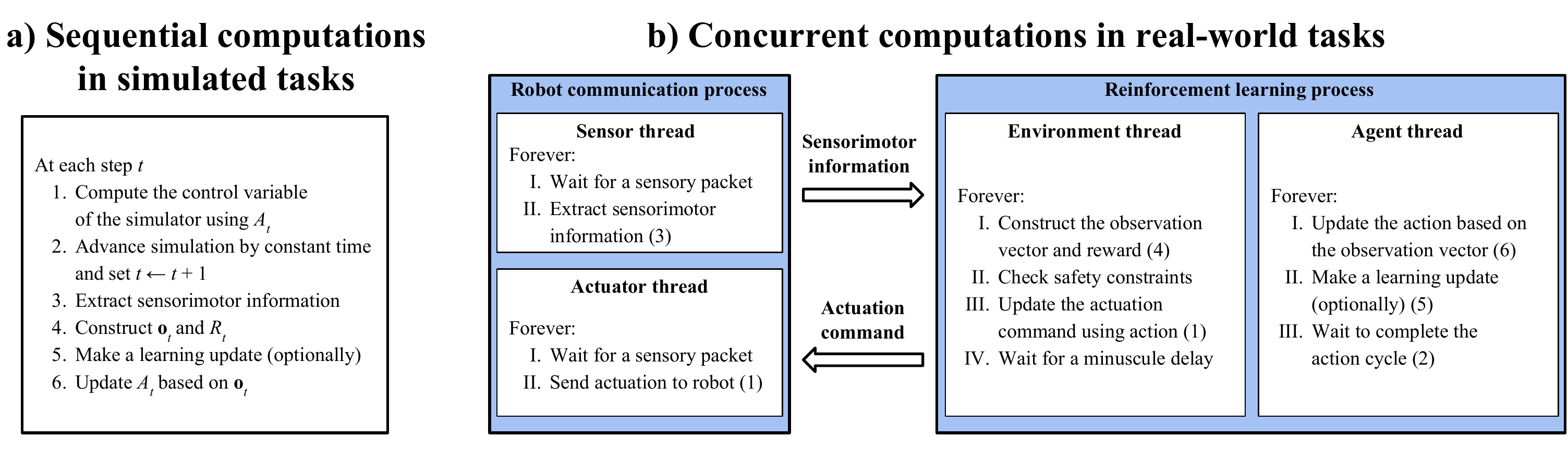}
\caption{
a) In typical simulated learning tasks, the world stands still during computations, and they can be executed sequentially without any consequence in learning performance.
b) The real world moves on during all computations and executing them concurrently can be desirable for minimizing delays.
}
\label{fig:computational-steps}
\end{figure*}

There are several other crucial aspects of a real-world task that are rarely studied in simulations, such as the action cycle time, the medium of connection, the choice of actuation type, and concurrency and delays in computation.
These aspects are the main focus of the current work.

\section{Elements of a Real-World Task Setup}

In this section, we describe the key elements of setting up a real-world learning task, different choices for each element, and the choice we make for our baseline UR5-Reacher setup.

\subsection{Concurrency, ordering and delays of computations}

In simulated tasks, it is natural to perform agent and environment-related computations synchronously, which may not be desirable in real-world tasks.
Figure 2(a) shows the computational steps that are  executed sequentially in a typical simulated experiment at each episode. 
The first four (1-4) computational steps are environment related whereas the last two (5, 6) are  agent related. 
The simulated world advances discretely in time during Step 2 and does not change during the rest. 
This way, simulated tasks can comply with the MDP framework, in which time does not advance between observing and acting.

In real-world tasks, time marches on during each agent and environment-related computations. 
Therefore, the agent always operates on delayed sensorimotor information. 
The overall latency can be further amplified by misplaced synchronization and ordering of computations, which may result in a more difficult learning problem and reduced potential for responsive control.
Therefore, a design objective in setting up a learning task is to manage and minimize delays.
Different approaches are proposed to alleviate this issue (Katsikopoulos \& Engelbrecht 2003, Walsh et al.\ 2009), such as augmenting the state space with actions or predicting the future state of action execution.
These approaches do not minimize the delay but compensate for it from the perspective of learning agents. 
A seldom discussed aspect of this issue is that different orderings or concurrencies of task computations may have different overall latencies.

In UR5 Reacher, we implemented the computational steps in Python and distributed them into two asynchronous processes: the \emph{robot communication process} and the \emph{reinforcement learning (RL) process}.
They exchange sensorimotor information and actuation commands. 
Figure 2(b) depicts the computational model of UR5 Reacher, which may also serve as a computational model for other real-world tasks.
Some of the computational steps in 2(b) end with step numbers from 2(a) when they are directly relevant. 
The robot communication process is a device driver which collects sensorimotor data from URControl in a \emph{sensor thread} at 8ms cycle time and sends actuation commands to it in a separate \emph{actuator thread}.
The RL process contains an \emph{environment thread} that checks spatial boundaries, computes the observation vector and the reward function based on UR5 sensorimotor packets and updates the actuation command for the actuator thread based on actions in a fast loop.
The \emph{agent thread} in the RL process defines task time steps and determines the action cycle time. 
It makes learning updates and computes actions using the agent's policy pass.

For the learning agent, we use the rllab implementation of TRPO. 
It performs computationally expensive learning updates infrequently, once every few episodes. 
We scheduled these updates between episodes to ensure that they do not interfere with the normal course of the agent's sensorimotor experience. 
Thus, learning updates of TRPO occur in the agent thread but not every action cycle or time step.

Our computational model of real-world tasks in Figure 5(b) suggests concurrency and certain ordering of computations to avoid unnecessary system delays. 
For example, splitting the robot communication process into two threads allows asynchronous communication with physical devices. 
Splitting the RL process into two threads allows checking safety constraints faster than and concurrently with action updates.
Moreover, we suggest making learning updates after updating the action, unlike Step 5 and 6 of simulated tasks (Figure 5a) where they are computed in the opposite order. 
This helps to dispatch actions as soon as they are computed instead of waiting for learning updates, which may increase observation-to-action delays. 
This computational model also extends to robotic tasks comprising multiple devices  by having one robot communication process per device, which allows the agent to access sensorimotor information fast. 

\subsection{The medium of data transmission}

It is natural to consider pairing a mobile robot with limited onboard computing power with a more computationally powerful base station via Wi-Fi or Bluetooth rather than USB or Ethernet.
Wi-Fi commonly introduces variability in the inter-arrival time of streamed packets.
In UR5 Reacher, the robot communication process communicates with URControl over a TCP/IP connection. 
We use an Ethernet connection for our baseline setup as communicating over Ethernet allows tighter control of system latency.
However, we also test the effect of using a Wi-Fi connection.
Figure 3 shows the variability in packet inter-arrival times for both wired and wireless connections measured using 10,000 packets. 
Packets are sent once every 8ms by URControl.
The Inter-arrival time is consistently around 8ms for the wired connection with all times between $[7.8, 8.6]$ ms.
For the wireless connection, there is much more variability with the complete range varying between $[0.2, 127]$ ms.

\begin{figure}
\center
\includegraphics[scale=.5]{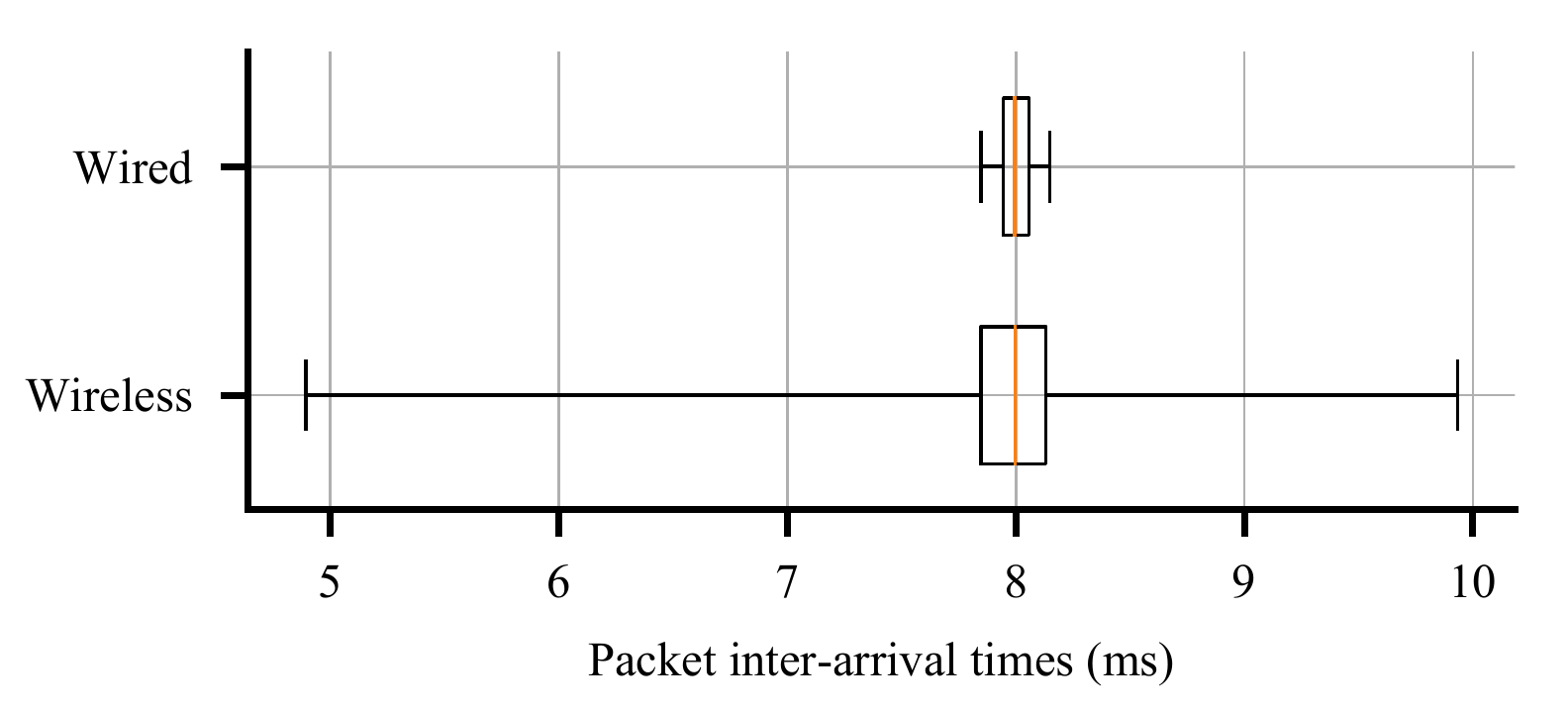}
\caption{
Packet inter-arrival times for the wired and the wireless connections used with UR5. The box plots show different percentiles (5, 25, 50, 75, 95).
}
\label{fig:computational-steps}
\end{figure}

\subsection{The rate of sending actuation commands to the Robot}

Different robotic devices operate in different ways. 
Some robots allow external computers to write directly in its control table.
The robot controller controls the actuators based on the control table and does not wait for instructions from the external computer.
Some other robots such as UR5 provide an interface where the controller controls the actuators based on actuation commands repeatedly sent by an external computer.
We refer to the transmission of these commands from an external computer to the robot as \emph{robot actuations}.
In UR5 Reacher, we choose the robot-actuation cycle time to be the default of 8ms.

\subsection{The action cycle time}

The action cycle time, also known as the \emph{time-step duration}, is the time between two subsequent action updates by the agent's policy.
Choosing a cycle time for a particular task is not obvious, and the literature lacks guidelines or investigations of this task-setup element.
Shorter cycle times may include superior policies with finer control.
However, if changes in subsequent observation vectors with too short cycle times are not perceptible to the agent, the result is a learning problem that is hard or impossible for existing learning methods.
Long cycle times may limit the set of possible policies and the precision of control but may also make the learning problem easier. 
If the cycle time is prolonged too much, it may also start to impede learning rate by slowing down the data-collection rate.

In our concurrent computational model, it is possible to choose action cycle times that are different than the robot-actuation cycle time.
When the action cycle time is longer, the actuator thread repeats sending the same command to the robot until a new command is computed based on a new action.
This may fortuitously benefit agents with long action cycle times. 
For example, to reach the target quickly, an arm may gain momentum by repeating similar robot actuations many times. 
Agents with a short cycle time must learn to gain such a momentum, while agents with a long cycle time get it for free by design.
We choose 40ms action cycle time in our baseline setup and compare the effects of both shorter and longer cycle times.

\subsection{The action space: position vs velocity control}

\begin{figure}
\center
\includegraphics[scale=.435]{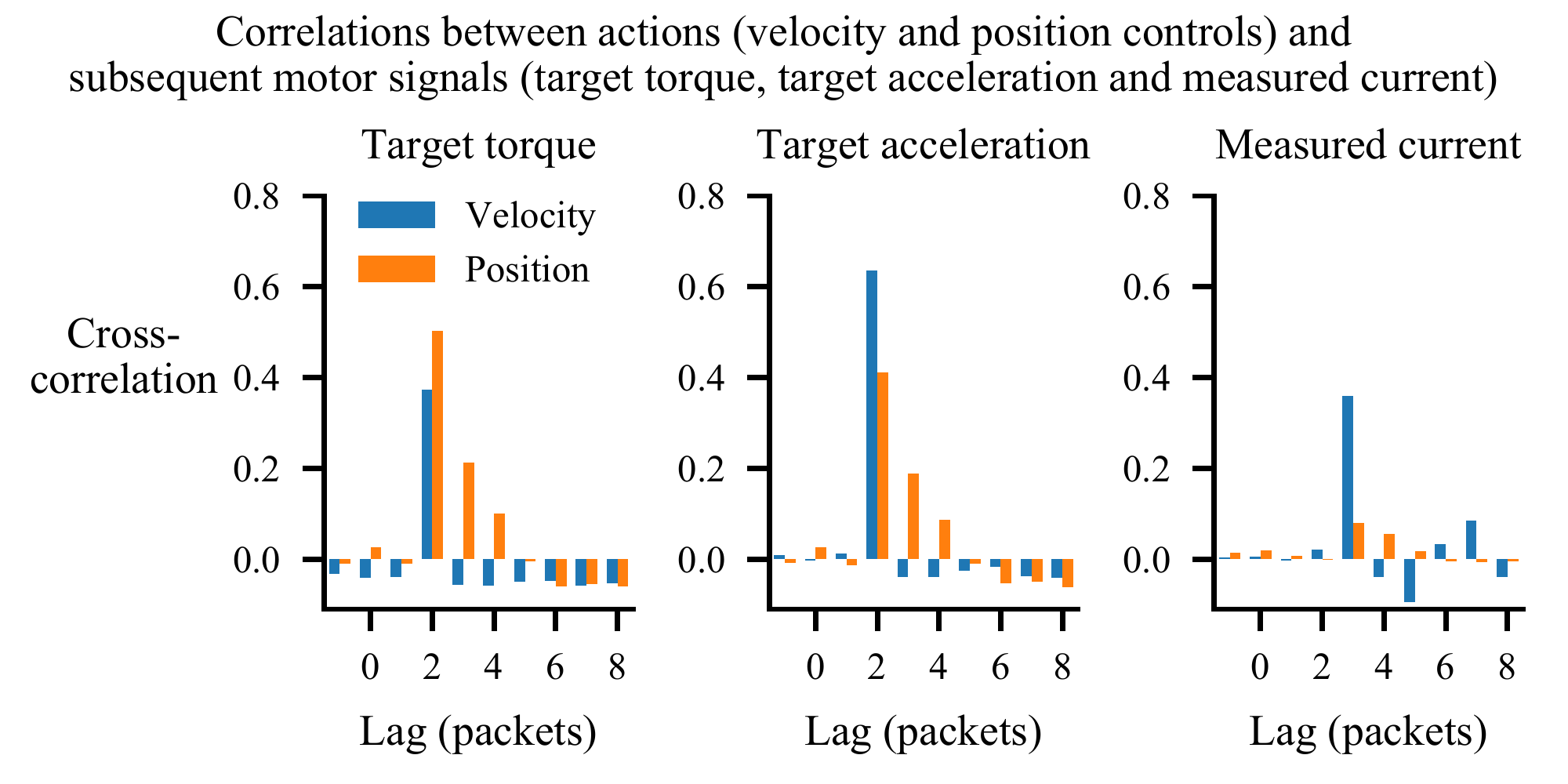}
\caption{
We compare actions based on velocity and position controls by showing their cross-correlations with three motor signals of UR5: target torque (\emph{left}), target acceleration (\emph{middle}), and measured current (\emph{right}). 
}
\label{fig:cross-correlations}
\end{figure}

\begin{figure*}
\center
\includegraphics[scale=.294]{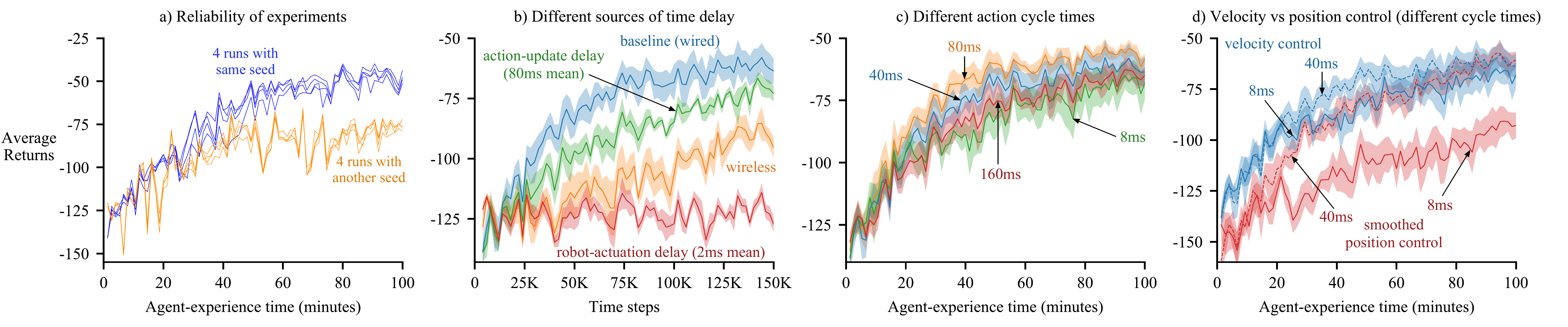}
\caption{
Learning performance in different UR5 Reacher setups.
{\bf a)} Our baseline setup allowed highly repeatable experiments with effective learning performance, where independent runs with same seeds provided similar learning curves. 
{\bf b)} Using a wireless connection or applying different sources of artificial delays were detrimental to learning in different degrees. 
{\bf c)} Using too short or long action cycle times resulted in worse learning performance compared to our baseline setup (40ms).
{\bf d)} Velocity control outperformed smoothed position control significantly with two different action cycle times.
}
\label{fig:learning-curves}
\end{figure*}

Choosing the action space can be difficult for real-world tasks as physical-robot controllers are usually not designed with the learning of low-level control in mind.
A control variable which has a strong cause-and-effect relationship with the robot's state in the immediate future is an appropriate choice for actions.
Torques or accelerations are often chosen as actions in simulated robot tasks.
We are interested in using a low-level actuation command for controlling UR5 to make the task similar to Gym Reacher. UR5 allows both position and velocity controls by sending commands every 8ms. 
We were able to use velocity control directly as actions, but using position control directly was not feasible.
The randomly-generated initial policies with direct position control generated sequences of angular position commands that caused violent, abrupt movements and emergency stops.

We choose direct velocity control for our baseline setup and compare with a smoothed form of position control.
To avoid abrupt robot movements with direct velocity control, we only needed to restrict the angular speeds between $[-0.3, +0.3]$ rad/s and the leading-axis acceleration to a maximum of 1.4 rad/s$^2$.
With position control, we needed to apply smoothing twice, where actions become a proxy to the second derivative of desired positions; applying smoothing once could not avoid abrupt movements. 
Our smoothing technique can be described as follows:
\begin{align}
\yv_t 			&= \clip_{y_{\min}}^{y_{\max}}(\yv_{t-1} + \tau\zv_t ), \\
{\qv}_{t}^{des} 	&= \clip_{\qv_{\min}}^{\qv_{\max}}(\qv_t + \tau\yv_t ).
\end{align}
Here $t$ is the agent time step, $\zv$ is the action vector, $\qv$ is the measured joint positions, $\qv^{des}$ is the desired joint position sent as the position-control command, and $\yv$ is the first derivative variable.
The $\clip_{a}^b$ operator clips a value between $[a, b]$. We set $\tau$ to be the action cycle time, $y_{\max} = -y_{\min} = 1$, and $(\qv_{\min}, \qv_{\max})$ according to the angle safety boundary. 
We choose the default value for the gain of the position-control command according to the URScript API.

URControl further modulates both position and velocity commands before moving the robot. 
Figure 4 shows the cross-correlation between both action types and three different motor signals based on data collected by a random agent with 8ms action cycle time. 
These motor signals are target acceleration, target torque and measured current, which are more closely related to subsequent motor events than other signals in UR5.
Both actions had their highest correlations with motor signals after two packets for target acceleration and torque, and after three packets for measured current.
This observation has driven the choice of our baseline cycle time to be longer than 8ms.
Velocity control had a higher correlation with motor signals than position control except for target torque. 
All the correlations for velocity control were concentrated in a single time shift whereas correlations of position control linger for multiple consecutive time shifts, indicating a more indirect relationship.

\section{Impacts of Different Task-Setup Elements}

We make variations to our baseline task setup to investigate the impact of different elements. 
In each variation of task setups, we used TRPO with the same hyper-parameters: a discount factor of 0.995, a batch-size of 20 episodes, and a step size of 0.04, which had the best overall performance on five different robotic Gym tasks. 
The policy is defined by a normal distribution where the mean and the standard deviation are represented by neural networks.
Both policy and critic networks use two hidden layers of 64 nodes each.
For each experiment, we run five independent trials and observe average returns over time. 

Neural networks are notorious for their dependence of performance on initial weights. 
Recently, Henderson et al.\ (2017) reminded how easily wrong conclusions could be drawn from experiments in which deep reinforcement learning methods were not applied carefully. 
It was exemplified by showing that the same algorithm may appear to achieve significantly different performance if the experiment is repeated using different sets of randomization seeds.
Therefore, two different methods or setups can seem significantly different simply due to random chance ensuing from different pseudo-random number sequences between them.
We took extra caution in setting up our experiments to ensure that in each task-setup variation the same set of five initial networks and five sequences of target positions were used. 

To validate the correctness of our experimental setup, we repeated the baseline experiment four times, as shown in Figure 5(a) for two different seeds. 
Each trial consists of 150,000 time steps or 100 minutes of agent-experience time and about three hours of total real-time including resets.
Over time each learning curve improved significantly, and the agent achieved higher average returns resulting in an effective and consistent reaching behavior, as shown in the companion video\footnote{
\url
{https://youtu.be/ZVIxt2rt1_4}
}.
Notably, all learning curves were quite similar to each other for the same seed even though they were generated by running each trial for multiple hours on different days and physical UR5 units. 
This is a testament to the precision of UR5, the stability of TRPO, and the reliability of our experimental and task setups. 

Figure 5(b) shows the impact of using a wireless connection. 
The solid lines are average returns, and the shaded regions are standard errors.
The wireless connection resulted in significant deterioration of performance compared to our baseline setup with a wired connection, which can be ascribed to variabilities and delays in the arrival of both sensorimotor packets to the computer and actuation commands to URControl.
To study their impacts we injected artificial exponential random delays, which crudely modeled Wi-Fi transmission delays, separately in action updates and the sending of actuation commands to URControl.
Action-update delays are in effect similar to observation delays (Katsikopoulos \& Engelbrecht 2003) and can also be caused by inefficient implementations.
Both delays can make a learning problem difficult by adding uncertainty in how actions affect subsequent observations whereas delaying robot actuations may additionally affect the robot's operation.
Figure 5(b) shows that a random action-update delay of mean 80ms caused significant deterioration in learning performance. 
On the other hand, adding a small random robot-actuation delay of mean 2ms devastated learning completely.

In Figure 5(c), we show the impact of choosing different action cycle times. 
The performance deteriorated when the cycle time was decreased to 8ms.
On the other hand, the performance improved when it was increased to 80ms, but deteriorated significantly from there when increased to 160ms.
In Figure 5(d), we show learning performance of both direct velocity and smoothed position controls for two different action cycle times: 8ms and 40ms. Smoothed position control performed \mbox{significantly} worse than velocity control in both cases.

\begin{figure}
\center
\includegraphics[scale=.45]{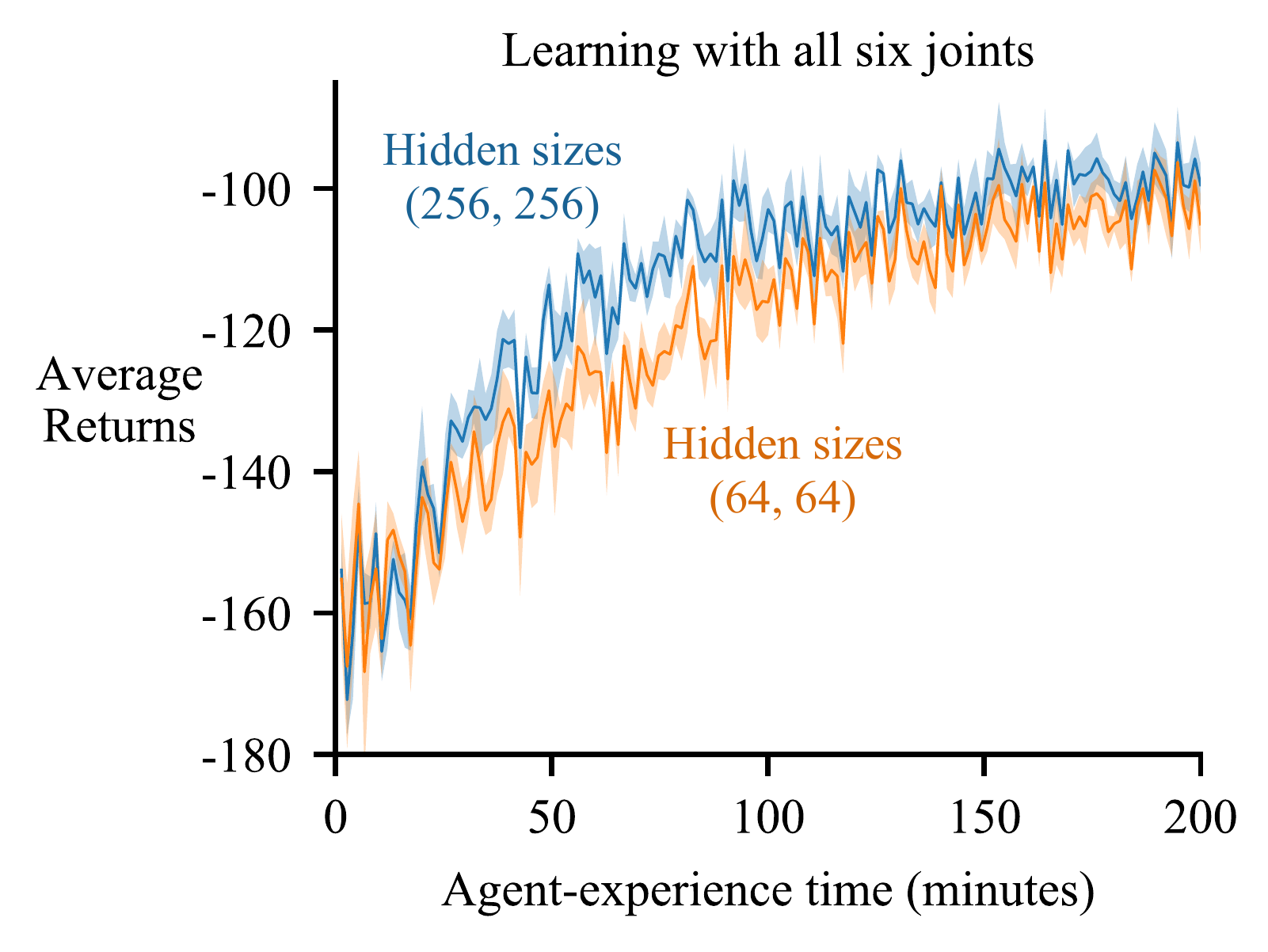}
\caption{Learning in UR5 Reacher 6D with all six joints actuated.}
\label{fig:6d}
\end{figure}

Finally, we investigated whether our baseline setup remained effective for six-joint control by applying it to UR5 Reacher 6D.
To accommodate the higher complexity of the problem, we explored policy and critic networks with larger hidden layers.
Figure 6 shows noticeable learning progress in 200 minutes, which continued to improve over time. The video shows effective behavior after running longer.

\section{Discussion and Conclusions}

In this work, we designed and developed a learning task step-by-step with a UR5 robot to discuss the key elements of real-world task setups.
Our discussion is summarized in the following hypotheses for real-world robotic learning tasks, which we used as a guideline for our baseline setup:
\begin{enumerate}[leftmargin=*]
\item 
System delays occurring in different computational stages are generally detrimental to learning.
Consequently, wired communications are preferable to the wireless ones.
\item 
Too small action cycle times make learning harder.
Too long action cycle times also impede performance as they reduce precision and cause slow data collection.
\item 
Choosing action spaces where actions are applied more directly to the robot makes learning easier by having more direct relationships with future observations.
\item 
Due to reduced delays, some concurrent computations are preferable to sequential computations of conventional simulated tasks.
\end{enumerate}
We studied the validity of the first three hypotheses by creating variations to our baseline setup, and our study largely supported them.
We demonstrated that learning performance could be highly sensitive to some setup elements, specifically, system delays and the choice of action spaces.
The performance was less sensitive to different action cycle times in comparison.
Our results suggest that mitigating delays from any source is likely beneficial, indicating the prospect of the last hypothesis.
Our study comprises only a small step toward a comprehensive understanding of real-world learning tasks, which requires more thorough investigations and validations using different tasks, learning methods and sets of hyper-parameters.
Our baseline setup allowed us to conduct highly reliable and repeatable real-world experiments using an off-the-shelf RL method. This served as a strong testament to the viability of extensive RL experimentations and research with real-world robots despite barriers and frustrations around robots and reproducibility of deep RL research, in general.

\section*{Acknowledgements}
We thank Richard Sutton, Lavi Shpigelman, Joseph \mbox{Modayil} and Cindy Yeh for their thoughtful suggestions and feedback that helped improve the manuscript.
We also thank William Ma and Gautham Vasan for recording and editing the companion video.

\section*{References}
\newcommand{\hangin}{\goodbreak\hangindent=.15cm \noindent}

\hangin
Brockman, G., Cheung, V., Pettersson, L., Schneider, J., Schulman, J., Tang, J., Zaremba, W.\ (2016). Openai gym. \emph{arXiv preprint} arXiv:1606.01540.

\hangin
Degris, T., Pilarski, P.\! M., Sutton, R.\! S.\! (2012).\! Model-free reinforcement learning with continuous action in practice. In \emph{American Control Conference}, pp: 2177--2182.

\hangin
Duan, Y., Chen, X., Houthooft, R., Schulman, J., Abbeel, P.\ (2016). Benchmarking deep reinforcement learning for continuous control. In \emph{International Conference on Machine Learning}, pp: 1329--1338.

\hangin
Gu, S., Holly, E., Lillicrap, T., Levine, S.\ (2017). Deep reinforcement learning for robotic manipulation with asynchronous off-policy updates. In \emph{IEEE International Conference on Robotics and Automation}, pp:3389--3396.

\hangin
Henderson, P., Islam, R., Bachman, P., Pineau, J., Precup, D., Meger, D. (2017). Deep reinforcement learning that matters. \emph{arXiv preprint} arXiv:1709.06560.

\hangin
Hester, T., Stone, P.~(2013). TEXPLORE: real-time sample-efficient reinforcement learning for robots. \emph{Machine learning 90}(3): 385--429.

\hangin
Katsikopoulos,\! K.\! V., Engelbrecht,\! S.\!~E.\!\! (2003).\! Markov\! decision processes\! with delays and asynchronous cost collection. \emph{IEEE transactions on automatic\! control 48}(4): 568--574.

\hangin
Levine, S., Finn, C., Darrell, T., Abbeel, P.~(2016). End-to-end training of deep visuomotor policies. \emph{The Journal of Machine Learning Research 17}(1): 1334--1373.

\hangin
Rusu, A.\ A., Ve\u cer\' ik, M., Roth\" orl, T., Heess, N., Pascanu, R., Hadsell, R.\ (2017). Sim-to-Real robot learning from pixels with progressive nets. In \emph{Conference on Robot Learning}, pp: 262--270.

\hangin
Schuitema, E., Wisse, M., Ramakers, T., Jonker, P.\ (2010). The design of LEO: a 2D bipedal walking robot for online autonomous reinforcement learning. In \emph{2010 IEEE/RSJ International Conference on Intelligent Robots and Systems}, pp: 3238--3243.

\hangin
Schulman, J., Levine, S., Abbeel, P., Jordan, M., Moritz, P.\ (2015). Trust region policy optimization. In \emph{International Conference on Machine Learning}, pp:1889--1897.

\hangin
Sutton, R.\ S., Barto, A.\ G.\ (2017). \emph{Reinforcement Learning: An Introduction (2nd Ed., in preparation)}. MIT Press.

\hangin
Walsh, T.\ J., Nouri, A., Li, L., Littman, M.\ L.\ (2009). Learning and planning in environments with delayed feedback. \emph{Autonomous Agents and Multi-Agent Systems  18}.

\hangin
Yahya, A., Li, A., Kalakrishnan, M., Chebotar, Y., Levine, S. (2017). Collective robot reinforcement learning with distributed asynchronous guided policy search. In \emph{2017 IEEE/RSJ International Conference on Intelligent Robots and Systems}, pp: 79--86.

\end{document}